\documentclass[journal]{IEEEtran}
\usepackage{subfigure}
\usepackage{cite}
\usepackage{hyperref}
\usepackage{csquotes}

\usepackage{balance}
\usepackage{color}
\usepackage{subfigure}
\usepackage{graphicx}
\usepackage{amsmath}
\usepackage{lipsum}

\usepackage{subfigure}
\usepackage{cite}
\usepackage{hyperref}
\usepackage{csquotes}
\usepackage{balance}
\usepackage{color}
\usepackage{subfigure}
\usepackage{graphicx}
\usepackage{tabularx}
\usepackage{lipsum}

\begin{document}

{\title{Towards Energy Efficient Distributed Federated Learning for 6G Networks}}
\author{Sunder Ali Khowaja$^\dagger$, \IEEEmembership{Member IEEE}, Kapal Dev*$^\dagger$\thanks{$^\dagger$Joint first authors, with equal contributions to this paper}, \IEEEmembership{Member IEEE}, Parus Khowaja, and Paolo Bellavista, \IEEEmembership{Senior Member IEEE}
\thanks{*Corresponding author}
\thanks{Sunder Ali Khowaja with Faculty of Engineering and Technology, University of Sindh, Jamshoro, Pakistan and Department of Mechatronics Engineering, Korea Polytechnic University, Republic of Korea. Email: sandar.ali@usindh.edu.pk, sunderali@kpu.ac.kr}
\thanks{Parus Khowaja is with University of Sindh, Jamshoro. (e-mail:Parus.khuwaja@usindh.edu.pk).}
\thanks{Kapal Dev is associated with the Department of Institute of Intelligent Systems, University of Johannesburg, South Africa, e-mail: (kapal.dev@ieee.org).}
\thanks{Paolo Bellavista is associated with University of Bologna, Italy. e-mail: (paolo.bellavista@unibo.it).}
}

%


\maketitle

\begin{abstract}
 	
The provision of communication services via portable and mobile devices, such as aerial base stations, is a crucial concept to be realized in 5G/6G networks. Conventionally, IoT/edge devices need to transmit the data directly to the base station for training the model using machine learning techniques. The data transmission introduces privacy issues that might lead to security concerns and monetary losses. Recently, Federated learning was proposed to partially solve privacy issues via model-sharing with base station. However, the centralized nature of federated learning only allow the devices within the vicinity of base stations to share the trained models. Furthermore, the long-range communication compels the devices to increase transmission power, which raises the energy efficiency concerns. In this work, we propose distributed federated learning (DBFL) framework that overcomes the connectivity and energy efficiency issues for distant devices. The DBFL framework is compatible with mobile edge computing architecture that connects the devices in a distributed manner using clustering protocols. Experimental results show that the framework increases the classification performance by 7.4\% in comparison to conventional federated learning while reducing the energy consumption. 
\end{abstract} 

\section{Introduction}\label{sec:intro}
Recently, a fifth-generation communication system (5G) has been rolled out to provide enhanced communication services in many developed countries. However, it is envisioned that 5G would suffer from a bottleneck when dealing with the increased number of communication embedded devices and applications demanding high bandwidth such as mixed reality, smart grid 2.0, unmanned aerial vehicles (UAVs), and industrial Internet of Things (IIoT) \cite{Jian2021}. Moreover, services such as enhanced energy efficiency, smart environments, and high-precision manufacturing in Industry 5.0 will not be effectively handled by the 5G communication system as they require sophisticated control, sensing, and computing functionalities \cite{Zhang2020}.
To facilitate the above-mentioned requirements, the sixth-generation (6G) communication system is in works that will be an ultra-dense network, extremely homogeneous, highly dynamic, and innately intelligent. To be specific, the 6G network is designed to operate at blistering data rate, i.e. multi-Tbps along with ultra-low latency \cite{Zhang2020}.
With the emergence of 6G communications, the provision of services to heterogeneous and massive machine-type communications (mMTC) devices has been gaining a lot of interest. The use of space-air-terrestrial-sea in 6G ensures service provisioning to the devices where communication infrastructure is not available \cite{Qi2020} by deploying a UAV as an aerial base station \cite{Niknam2020}. The UAVs can also be used as mobile relays or data collection units, respectively. As suggested earlier, 6G needs to accommodate mMTC devices with ultra-low latency, therefore, AI assisted techniques, specifically machine learning algorithms, need to be integrated in UAVs to provide effective service.
Recently, various strategies to train machine learning models have been proposed such as transfer learning, active learning, and federated learning. The federated learning approach has garnered a lot of interest due to its co-operative training approach and security characteristics. In principle, the technique acquires a trained machine learning model using the private data on local devices or machines. The updates from the trained models are sent to the UAV serving as an aerial base station without altering or accessing the private data. The UAV then aggregates all local updates from the devices to create a global model and broadcasts it to all the devices within its coverage area. The process continues till the convergence of global model with an adequate test accuracy \cite{Samarakoon2020}. The whole process can better preserve the security and privacy of data with evolving cyber attacks on the horizon \cite{KhowajaQ2021}. Existing studies suggest that the federated learning approach is also effective for reducing the traffic load, computational overhead, and latency \cite{Yu2021}. It is apparent that the federated learning provides an edge in comparison to transfer learning and active learning in the context of 6G communication systems.
\begin{figure*}[h]
  \includegraphics[width=\linewidth]{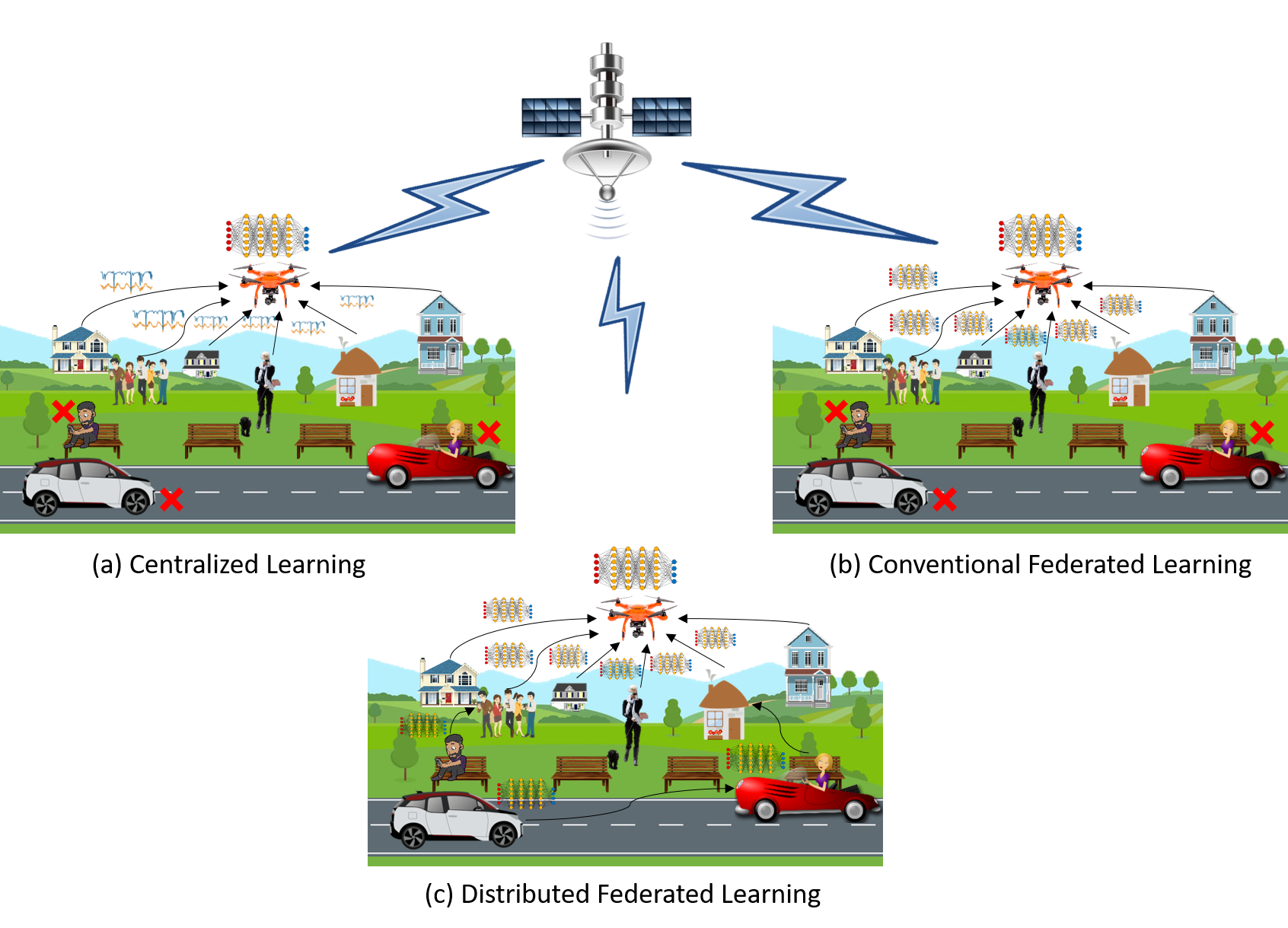}
  {\caption{Visual Comparison of Centralized and Conventional Federated Learning with Distributed Federated learning for Futuristic Networks}}
  \label{Fig1}
 
\end{figure*}
However, due to the continuous communication with a stationary or aerial base station, the problem of energy consumption concerning sustainable wireless infrastructure still persists \cite{Salameh2021}. Note that the problem stands for both, the users in the coverage area and the base station itself, since they are battery or electricity powered devices \cite{Bian2020}. To overcome the aforementioned issues, we propose the use of distributed federated learning which corresponds to a combination of federated and distributed learning \cite{Elgabli2020} that enables the local clients to update their trained model without being directly connected to the corresponding UAV. Similarly, the UAV does not have to update all the clients within its range with the global model. The UAVs can send the updated global model to nearby devices, which then propagate the model further to other clients via device-to-device communication. Such type of communication not only preserves data privacy but also helps in providing long-range communication, accommodate massive machine-type communication devices, and helps in designing sustainable energy infrastructures. To the best of our knowledge, a detailed architecture for a distributed federated learning approach in compliance with mobile edge computing framework has not been proposed yet. The contributions of this work are as follows:
\begin{itemize}
    \item We propose a distributed federated learning based framework for energy efficient communication in futuristic networks.
    \item We show that the proposed framework can handle heterogeneous devices and labels for detection tasks.
    \item We show that the proposed network improves the energy efficiency through experimental analysis.
    \end{itemize}
The rest of the paper is structured as follows: Section 2 provides an overview of the conventional and distributed federated learning. Section
3 presents the proposed DBFL framework. Section 4 discusses the handling of heterogeneous feature space. Section 5 presents the preliminary analysis and its comparison with the conventional federated learning approach. Section 6 concludes the study and highlights some future research directions.
\begin{figure*}[h]
\centering
  \includegraphics[width=\linewidth]{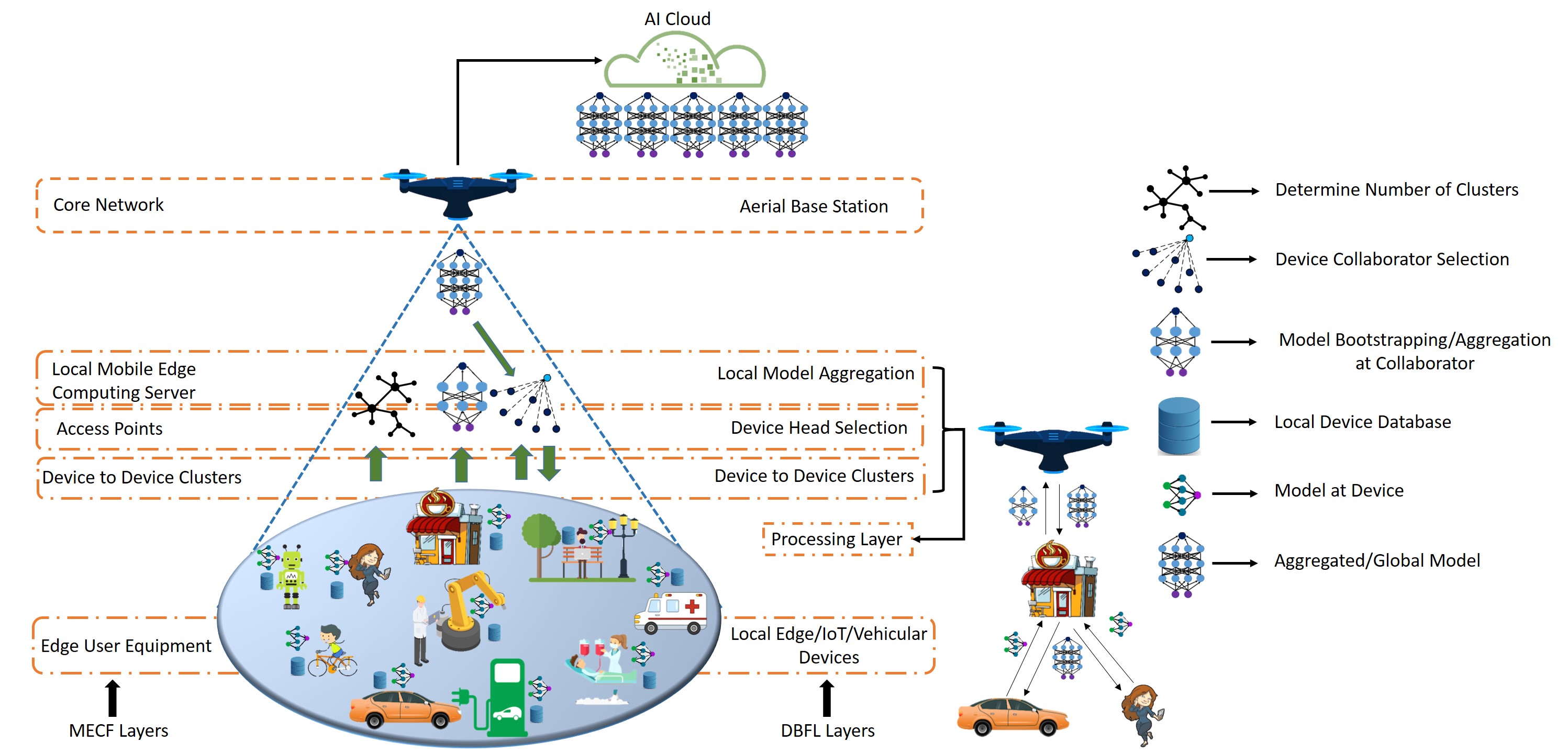}
  \caption{Proposed Distributed Federated Learning. The layers on the left side correspond to the compliance with mobile edge
computing framework (MECF) while the layers on the right represent DBL framework}
  \label{Fig1}
\end{figure*}

\section{Conventional and Distributed Federated Learning}
The federated learning approach was originally proposed to cope with the centralized learning system that collects the data from local clients, trains the machine learning model, and transmits it to the devices for further usage. The main advantage of the centralized learning approach is that the wireless network does not affect the model performance since it is trained at the base station. However, the centralized approach is not secure as it is prone to data leakage. Furthermore, the devices and network need to add significant communication resources and overhead (due to the direct transmission of data) to execute centralized learning \cite{Chen2020}. \\
The conventional federated learning (CVFL) framework was first proposed by Google that allows the machine learning model to be trained in a cooperative manner without sharing the data \cite{hard2018federated}. The CVFL assumes that the employed machine learning architecture for both the local client and the aerial base station is the same. The CVFL approach takes into account the data available to the edge device and trains the model locally. All the locally trained models are then sent to the aerial base station for further aggregation. This process is performed in an iterative manner until the convergence. The local device models are then updated with the aggregated (global model) once it converges \cite{Ye2020}. The main advantage of the CVFL is the privacy of data that is preserved when training and updating the machine learning model.
In comparison to the centralized learning, CVFL also uses less computational overhead. CVFL disadvantages include the detection performance that is affected by the wireless transmission, increased
convergence time, reduced energy efficiency, and at times unavailability of communication \cite{Chen2020}. \\
In practical communication systems, all devices are unable to connect to an aerial base station potentially due to transmission delays and energy constraints. To cope with this issue, distributed federated learning (DBFL) is proposed that can participate in model convergence as well as receive the aggregated model without being connected to the aerial base station. As the name suggests, the DBFL works on the principle of multinode communication where the local clients can associate with their neighboring nodes for indirect connection to the aerial base station. The idea behind DBFL is that, if the local clients are not able to directly connect to the aerial base station, they can still perform the federated learning through their neighboring devices selected through nearest neighbor or cluster head selection algorithms. The iterative training process and transmission of aggregated models will also be performed in a distributed manner. The DBFL reduces the energy consumption and computational overhead by reducing the long-range communication and sharing the computational load with
other devices, respectively. This reduction in consumption and computational load leads to energy efficient communication. Furthermore, the DBFL also cope with the cold-start problem considering that the newly registered device in the cell has not enough data to train the model locally. The model shared with DBFL can help in making the inferences at start followed by fine-tuning the already trained model. The CL, CVFL, and DBFL are compared visually in Figure 1 and the key differences are highlighted in Table 1.
\begin{table}[]
\caption{KEY DIFFERENCES BETWEEN CVFL AND DBFL}
\label{tab:my-table}
\begin{tabular}{|l|l|l|}
\hline
No & \multicolumn{1}{c|}{\textbf{CVFL}} & \multicolumn{1}{c|}{\textbf{DBFL}} \\ \hline
1 & \begin{tabular}[c]{@{}l@{}}All devices needs to have a \\ reliable and direct connection \\ to BS.\end{tabular} & \begin{tabular}[c]{@{}l@{}}Devices can either connect \\ directly to BS or to any other \\ device within the cluster having \\ a reliable connection.\end{tabular} \\ \hline
2 & \begin{tabular}[c]{@{}l@{}}Number of users in CVFL \\ are limited.\end{tabular} & \begin{tabular}[c]{@{}l@{}}Can accommodate more devices \\ in comparison to CVFL.\end{tabular} \\ \hline
3 & \begin{tabular}[c]{@{}l@{}}Only homogeneous data \\ and feature space are \\ supported.\end{tabular} & \begin{tabular}[c]{@{}l@{}}Supports homogeneous as well \\ as heterogeneous data and \\ feature space.\end{tabular} \\ \hline
4 & \begin{tabular}[c]{@{}l@{}}The support for energy \\ efficiency is limited.\end{tabular} & \begin{tabular}[c]{@{}l@{}}Is designed to improve the \\ energy efficiency.\end{tabular} \\ \hline
5 & \begin{tabular}[c]{@{}l@{}}Low computational \\ overhead.\end{tabular} & \begin{tabular}[c]{@{}l@{}}Slightly increased \\ computational complexity.\end{tabular} \\ \hline
\end{tabular}
\end{table}
\section{Distributed Federated Learning Framework}
As suggested earlier, 6G and futuristic networks demand drastic increase in network traffic which also results in increasing computational load on the aerial base stations. Only relying on the aerial base station would result in comprised quality of service (QoS). Additionally, the energy efficiency is also affected due to increased communication service. Therefore, it is a necessity to utilize the computation as well as communication resources in an effective manner that could be the basis of designing sustainable energy infrastructures. In this regard, we propose the distributed federated learning framework as shown in Figure 2. The framework proposes four layers, i.e., Local Edge/IoT/Vehicular device layer, Processing layer, Aerial base station layer, and AI cloud layer. Since the AI cloud layer aggregates the model in a same way as the Aerial base station layer, but on a larger scale, we discuss the first three layers in detail. The architecture is compliant with the Mobile Edge Computing framework (MECF) \cite{Zhang2020} that focuses on enhancing the communication and computing capability of the communication network. The DBFL layers (on the right) are mapped to the corresponding MECF layers (on the left) in Figure 2. 


\textit{The Local Edge/IoT/Vehicular layer:} This includes both stationary and mobile devices, accordingly. The framework assumes that each of the devices in this layer comprises of a local database and computational resources for storing the data and training the model locally. It is also assumed that each of the devices that trains the model has homogeneous data, feature space, and labels, respectively. The DBFL framework can also deal with heterogeneous data and feature space, but it requires additional processing. This layer is in compliance with the edge user equipment in MECF that is responsible for collecting the data and transferring it to the base station or the subsequent layer, accordingly. \\
\textit{Processing Layer:} The processing layer in DBFL framework maps to the mobile edge computing layer in MECF. We highlight the difference between DBFL and MECF with respect to each corresponding layer. The processing layer in DBFL is further divided into three processing units, i.e. \textit{Device-to-Device Clusters}, \textit{Device Head Selection}, and \textit{Local Model Aggregation.} The \textit{Device-to-Device Cluster (D2DC)} comprises of two tasks, the first is to find the devices within the pre-specified range to form a cluster and the second is to check the compatibility in terms of homogeneity of the data and labels. The inference rules, optimization algorithms and clustering techniques can be used for the formation of clusters based on a certain criterion. For instance, in the example shown in Figure 2, the criterion can be considered as follows:

\begin{itemize}
    \item A cluster should consist of three devices at most.
    \item A cluster should have at least one device that can directly be connected to aerial base station.
\end{itemize}

Further characterizations to the cluster formation can be added based on the domain application, or user preference. The second task which D2DC needs to perform is the homogeneity of the data and labels. For instance, the model of cars and the model of a bicycle have compatible feature space (same number of features and dimensions) but may differ in class labels, therefore, the homogeneity criterion is not fulfilled. Once the homogeneity is checked, the cluster formation can be performed recursively in case of the cluster devices are not compatible. The D2DC unit in MECF is also responsible for forming clusters based on similar characteristics, however, the formation of clusters is rather hypothetical and can be modified, accordingly.\\
After the formation of clusters using D2DC, the \textit{Device Head Selection} (DHS) needs to be performed. It is similar to the cluster head selection techniques where a cluster head is selected and is held responsible for all communications. However, considering the energy and mobility constraints, further inference rules or optimization algorithms can be opted for selecting the cluster head. An example of inference rules is given below:

\begin{itemize}
    \item A device head should be capable of direct communication to the aerial base station
    \item A device head should have lower aggregated distance from the devices within a cluster
    \item In case of the same distance, the device head should be preferred with respect to battery life
    \item In case of same battery life, a stationary device head should be preferred
    \item In case all the devices are stationary, the device head with lower latency to the aerial base station should be preferred
\end{itemize}

The DHS should typically be performed after every 2-5 minutes or lower considering the mobility and energy constraints. In comparison to MECF, the DHS combines some of the characteristics of D2DC and Local Mobile Edge Computing Server (LMECS). The MECF presumes that the devices are capable of communicating and processing data in an autonomous manner. The assumption that all the devices are of same standard and computation capabilities is far from realization. The DHS in DBFL adds some constraints while selecting a device so that the computation is handled in an effective and energy efficient manner.  
The last processing unit is the \textit{Local Model Aggregation} (LMA) which will be aggregated by the device head. Considering that the data, feature, and label space are homogeneous, the aggregation can be performed by either model weighted averaging, model adaptive weighted averaging,
meta-learning, or simply retraining. The selection of the aggregation method depends on the availability of computation and energy resources at the end of the device head. This selection can be performed either based on an optimization algorithm or static rule based inference engine. The models from devices in a cluster are sent to the device head which can perform any of the aforementioned methods to aggregate the trained models. The weighted averaging simply uses static weights for averaging the class probabilities and selects the model closest to the aggregated one \cite{Khowaja20}. The adaptive weighted averaging optimizes the weight of each corresponding class label before averaging the class probabilities. Once averaged, the model closest to the aggregated one is selected for further propagation \cite{Khowaja20}. The meta-learning technique trains an individual shallow classifier on the output class probabilities of the obtained models. It has been proven by the existing studies that meta-learning approach achieves better results than the weighted and adaptive weighted averaging method \cite{Khowaja20}. However, there is a trade-off when using meta-learning, i.e. user willing to share data (security) and sending an extra meta-classifier (computational and storage overhead). If the domain requires high accuracy and the data is not sensitive, then the meta-learning can be used; otherwise, the averaging method can be employed. A similar case can be made for re-training that requires not just labels but complete data along with the models. The LMA unit also combines the partial functionalities of LMECS and the core network in MECF. The model aggregation process in MECF is performed at the core network stage, however, it requires the data to be sent directly or through edge servers to the base station. As discussed, the transmission of data raises security and privacy concerns, therefore, the LMA unit in DBFL performs the task locally on the device selected as the cluster head and sends the model to the aerial base station, accordingly, thus reducing the possibilities of security breach or data interception.\\
\textit{Aerial Base Station}: Further communication will be carried out between the device heads and the aerial base station. All device heads will send the aggregated models to the aerial base station for global model aggregation. The aerial base station can perform the model aggregation in a similar manner, i.e. weighted and adaptive weighted averaging schemes \cite{Khowaja20}. However, if provided with the corresponding labels, meta-learning approach can also be dealt at the aerial base station layer. Once the global model is aggregated, the model is then transmitted to the local edge/IoT/vehicular devices through the device heads. This process will not only reduce the communication processes but will also balance the computational overhead amongst all devices involved in the model update process, which eventually leads to increased energy efficiency. The core network layer in MECF and aerial base station in DBFL are almost compliant with the difference in scale of computations and processing. The core network in MECF can be regarded as a centralized approach shown in Figure 1 that only considers the devices within the available range, whereas the DBFL is decentralized, covers a larger area through D2DC extension, and relies on enhancing energy efficiency.\\
A hierarchical model updated in diversified manner can be performed using AI cloud which may collect the global aggregated models from multiple aerial base stations and update them to form a meta-global model that can be then shared to the local devices via aerial base station and the corresponding device head.

\section{Generalized Homogeneity using DBFL Framework}
We briefly discuss an example of model training in terms of homogeneous and heterogeneous feature space using DBFL framework. The CVFL assumes that the same model should be opted by all local devices as well as the base station to maintain the homogeneity of data, feature, and label space. However, the assumption of using artificial neural networks and deep learning approaches requires huge amount of data to train the model in an effective way. The shallow learning methods are still preferred to deal with limited amount of data, but those methods heavily rely on feature engineering techniques in order to attain higher levels of accuracy. Subsequently, the feature space may vary in terms of number of dimensions, characteristics, sampling rate, and so forth. This phenomenon is referred to as heterogeneous feature space that can create problems when sending the model to device heads and model aggregation. In this regard, we propose the use of autoencoders (AE) stacked with the features extracted for making the feature space homogeneous \cite{KhowajaQ2021}.\\
For the homogeneous part, the raw data at local database of user devices will undergo the training process using the artificial neural networks having same parameters. This results in a homogeneous feature space, thus does not create any problem during the transmission and aggregation of models. However, if some feature extraction techniques are used, such as principle component analysis, Fourier analysis, and more, the feature space varies vastly and a trained shallow learning model using different set of features can cause difficulty in aggregation. In this regard, we propose the transformation of feature space using AEs that constructs a latent space representation, which is then trained with the shallow classifier. Furthermore, existing studies have shown that the AEs can be generalized to multiple data modalities such as image, text, and speech, however, the layers, number of neurons, objective function, and parameters needs to be varied, modified and optimized, accordingly \cite{KhowajaQ2021}.
\begin{figure}[h]
\centering
  \includegraphics[width=\linewidth]{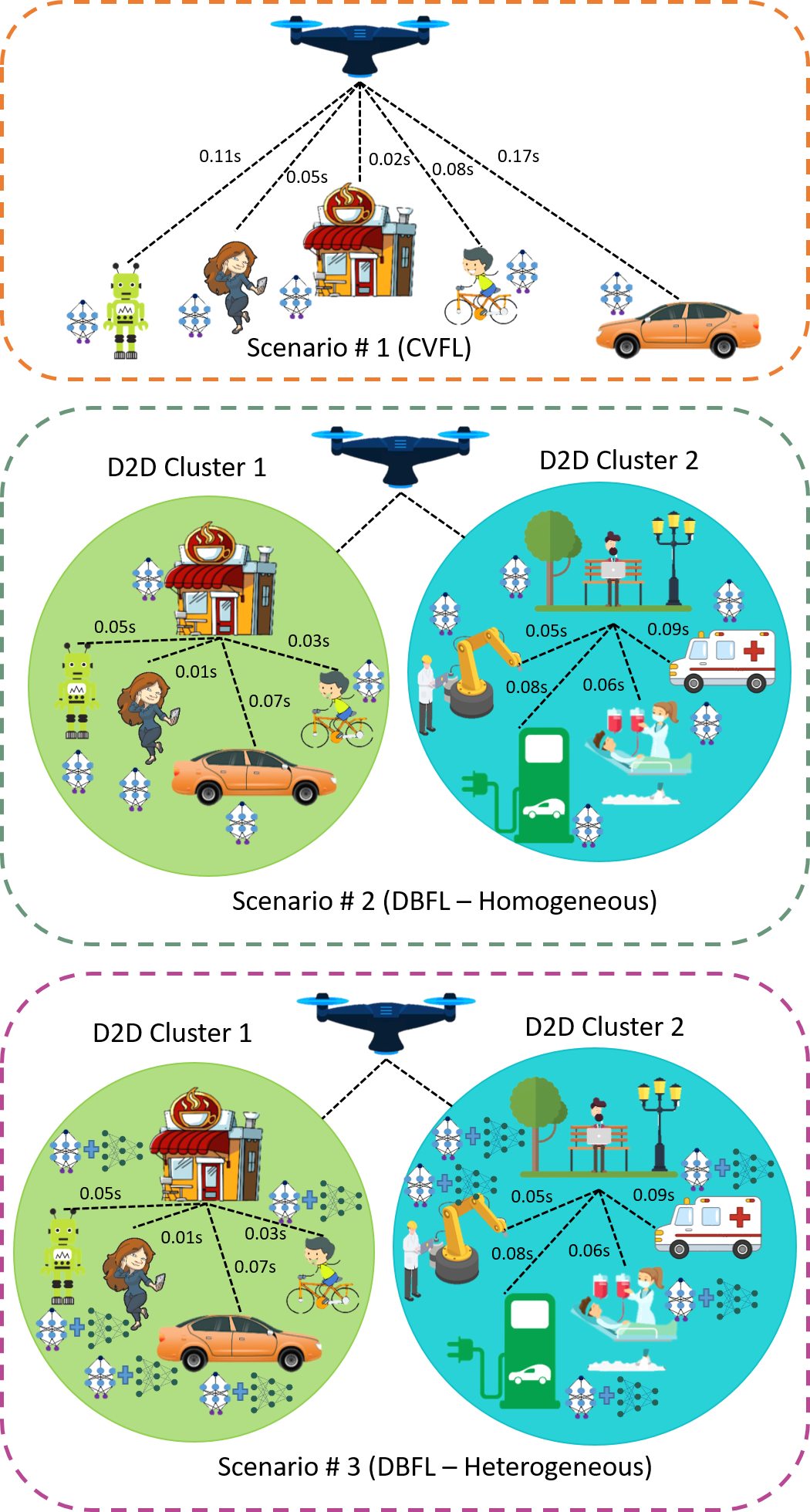}
  \caption{Illustration of Scenarios for Preliminary Analysis}
  \label{Fig4}
\end{figure}
\section{Experimental analysis}
We performed the analysis to compare the CVFL and DBFL approach in terms of accuracy and energy efficiency to show the proposed work’s effectiveness. We consider three scenarios as shown in Figure 3. All the scenarios consider one aerial base station and five devices in the coverage area. The number of devices were set to five (i) for dealing with a relatively large but manageable and easy-to-understand number of devices and (ii) for evenly dividing them into two clusters considering the example shown in Figure 2. The first scenario represents the CVFL approach where each of the devices in the area needs to connect to the aerial base station. However, it can be seen that 2 out of 5 devices in this scenario cannot connect to the aerial base station due to high latency issues. The second scenario represents the DBFL approach where devices can connect with a device head in a cluster which then sends the aggregated model to the aerial base station. This scenario uses a homogeneous feature space and for the sake of simplicity, we show different hypothetical D2D clusters without performing actual operations. The third scenario represents the DBFL approach with similar characteristics as the second scenario, but a heterogeneous feature space is used. We employ the IoT device type identification dataset proposed in \cite{meidan2017detection} that has 9 different types of IoT devices. The reason for the selection of the said dataset is two-fold. The first is the availability of the dataset in public domain and the second is the compliance of dataset characteristics with the domain of the proposed study, i.e., Futuristic Networks. Currently, 5G and 6G networks extensively use preamble detection and active user detection for new radios that somehow can bear a resemblance to the device type classification. Furthermore, in DBFL, the device type needs to be identified in order to select the device head, therefore, the dataset seems to be compliant and provides features for a wide variety of IoT devices or user equipment that can be used for 6G networks. \\
For homogeneous CVFL and DBFL part, it is assumed that each device is trained with artificial neural network having 80 neurons and 3500 data samples. The maximum transmission time is set to 0.1 seconds, beyond this time, the device cannot connect to either the aerial base station or the device head. The latency for each device was manually set to comply with the designed scenarios. The DBFL framework automatically chooses the weighted averaging for model aggregation by default due to its low computational complexity in comparison to other methods. The method averages the class probabilities to aggregate the model at the device head and aerial base station level.  \\
For heterogeneous DBFL, we assume that each device uses different dimensions of features but undergoes the AE to get the unified feature space dimension. For instance, the employed dataset has around 274 features. One device may use 60 features while the others might use number of features within the range of 1-274. These features will be the input to the AEs which generates, say, 15 dimensions of feature space. For the sake of simplicity, we chose the output of 25 while rest of the parameters are the same with the Scenario$\#2$, respectively.

The device type classification accuracies for all three scenarios over 100 iterations are shown in Figure 4 (a). It is evident from the analysis that the CVFL (Scenario$\#1$) gets stuck around 0.81-0.87 which is by assumption due to the fact that only two devices can connect to the aerial base station as the other two exceed the transmission time, thus, their connection to the base station is denied. Furthermore, the error bars indicate that the deviation in the accuracies are quite high for low data volume in the initial iterations. The DBFL-Homogeneous (Scenario$\#2$) and DBFL-Heterogeneous (Scenario$\#3$) performs better and converges faster than the CVFL. In Scenario$\#2$ it can be seen that the all the devices in a D2D cluster participate for data aggregation purposes, which increase the performance level in terms of accuracy and faster convergence. Furthermore, it should also be noted that the transmission delays are reduced due to the device head being closer than the aerial base station. Scenario$\#3$ slightly improves the accuracy and converges slightly faster. However, the improvement is recorded on the expense of added computation as each device has to transform the feature space using AEs followed by training and transmission. In terms of error bars, the deviations are almost or less than half in comparison to CVFL. Therefore, we have restricted the number of features to 50 for each device, however, there are more than 250 features available in the dataset. We presume that the accuracy can further be improved by adding more features at the cost of slightly increased computational complexity. 
\begin{figure*}[h]
\centering
  \includegraphics[width=\textwidth]{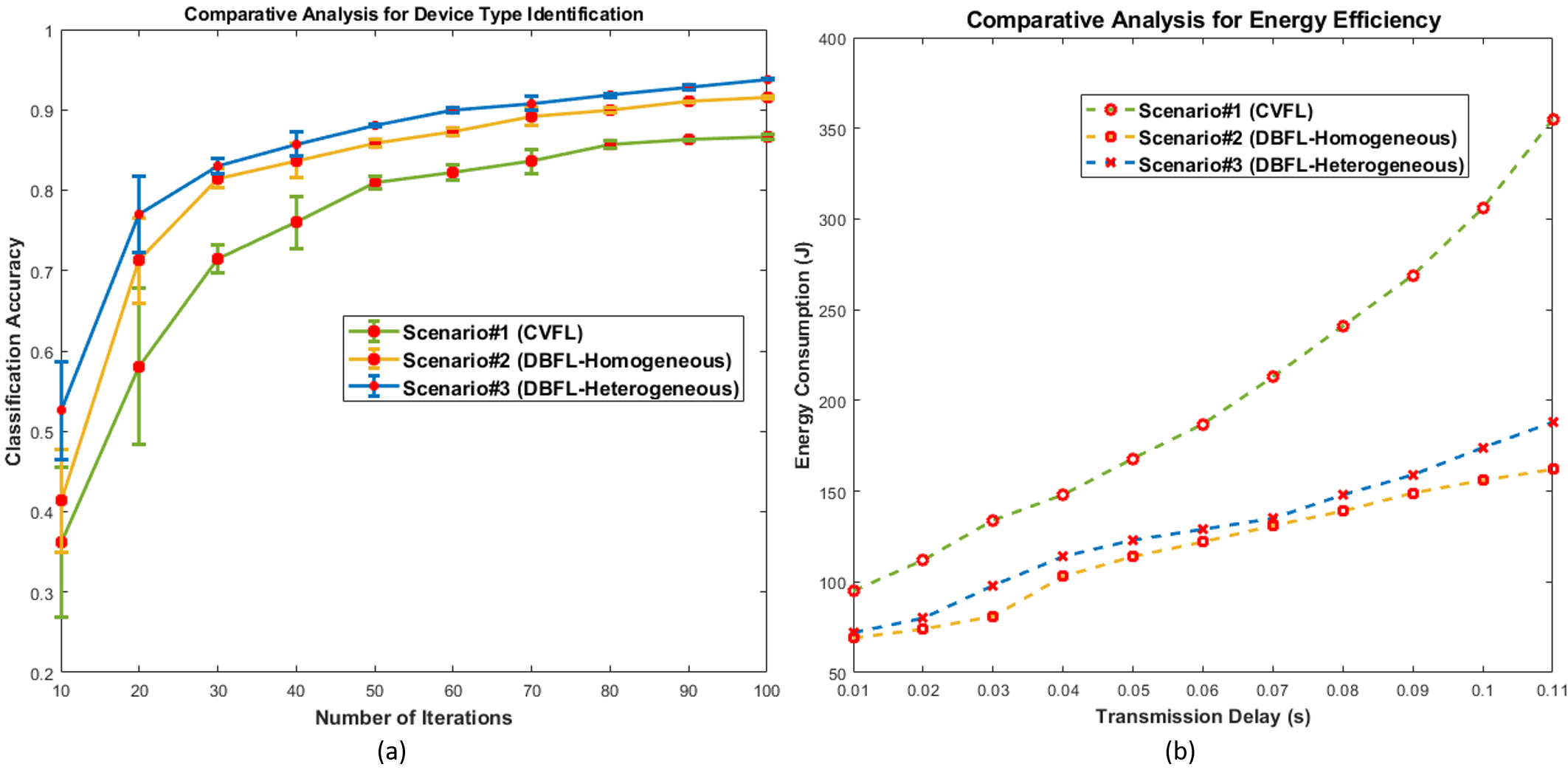}
  \caption{Comparative analysis for (a) IoT device type identification accuracy with multiple scenarios and (b) Energy Consumption while varying transmission delay with multiple scenarios}
  \label{Fig5}
\end{figure*}
\\
We also performed an experimental analysis on the energy consumption of local edge/IoT/vehicular devices while varying the transmission delay. A total of 5 nodes were created with three fixed and two mobile nodes. The attenuation factor was set to 2.0 and the energy values for the nodes were set between 80-100, where 100 was assigned to the device head. The consumption were cycle was varied between the values 0.2-0.35 with respect to the communication distance and computational overhead. The simulation has been carried out in MATLAB R2019b. The results for comparative analysis in terms of energy consumption is shown in Figure 4 (b). It is well established that increase in communication distance increases the transmission delay, which eventually decreases the energy efficiency. In Scenario$\#2$, it is shown that the transmission delay increases as the devices get farther and farther. This leads the devices to increase the transmission power for establishing the communication with the aerial base station. Hence, the energy consumption of the devices increases accordingly. In Scenario$\#2$, the transmission delay is overall decreased by adding a device head to D2D cluster. The effect decreases the transmission power and thus leads to an energy efficient communication. The DBFL-Heterogeneous slightly increases the energy consumption, which, we believe is due to the increased computational complexity.  


\section{DBFL Open Issues and Challenges}
The DBFL framework largely targets the networks 5G/6G and beyond, therefore, it is important to highlight the open issues and challenges that could help in improving the DBFL’s performance.\\
\textit{Objective function:} The objective function in federated learning can be customized based on the domain application. However, the computing resources also vary with respect to the employed objective function. The parameters such as number of iterations and model size also vary accordingly. It should be noted that the model size increases every time the model is aggregated. The convergence time is also related with the objective function. Therefore, the design of an objective function that is less computational complex and reduces the model size is of great importance.\\
\textit{Scalability and Aggregation: }The concept of massive machine-type communication suggests that a single aerial base station has to accommodate a large number of devices that open doors for vast research opportunities. As suggested in earlier sections, the method for selecting device heads on an alternative basis can also help in increasing energy efficiency and accommodating more number of devices. On the other hand, the model aggregation considers the weighted averaging method in a hierarchical manner, however, there are many sophisticated methods that could be used for the said purpose or combining the decision class probabilities for increasing the detection/classification performance. \\
\textit{Verification and Validation Measures:} Although this study performs the evaluation of said tasks based on accuracy, energy consumption, and transmission delay as suggested in existing works, the use of verification and validation measures remains an open issue. Furthermore, these measures vastly vary with respect to the application domain, type of system, and type of data, accordingly. For instance, the hard real-time systems should be validated in terms of computation time and complexity as the systems need to respond in matter of microseconds. Meanwhile, if system deals with medical reports, the validation measure should be accuracy and F1-score instead of complexity and run-time. Similarly, if systems deal with sensitive data and the priority is to secure it, then the validation measure needs to be in compliance with GDPR regulations. Considering the aforementioned facts, the selection of verification and validation measures still remains an open issue concerning DBFL.
\section{Conclusion}
In this work, we proposed the use of a distributed federated learning approach that not only resolves the issues associated with centralized systems but also exhibits scalable, faster, and energy-efficient communication when exchanging the trained models. This DBFL also caters to the data- and feature-space heterogeneity issues for model aggregation via autoencoders. We performed experiments to show that the DBFL-Homogeneous performs better in terms of accuracy as well as energy efficiency in comparison to the CVFL. We also provided some potential future directions that can be considered when carrying the DBFL framework forward for further testing and prototype implementation purposes.



\bibliography{ref.bib}
\bibliographystyle{IEEEtran}
\vskip -0.1\baselineskip plus -1fil
\vspace{-0.25cm}
\begin{IEEEbiographynophoto} {Dr. Sunder Ali Khowaja} {Dr. Sunder Ali Khowaja} received the Ph.D. degree in Industrial and Information Systems Engineering from Hankuk University of Foreign Studies, South Korea.  He has served as an Assistant Professor at Department of Telecommunication Engineering, University of Sindh, Pakistan. He is currently associated with Department of Mechatronics Engineering, Korea Polytechnic University, Republic of Korea, in the capacity of postdoctoral research fellow. He is also serving as a reviewer for many reputed journals, including, IEEE Transactions on Industrial Informatics, IEEE Access, IEEE Internet of Things Journal, IEEE Transactions on Network Science and Engineering, IEEE Transactions on Medical Imaging, and others. He also served as a Technical Program Committee member in CCNC 2021, Mobicom 2021, and Globecom 2021 workshops. He is currently assisting in the capacity of Guest Editor at Computers and Electrical Engineering, Human-Centric Computing and Information Sciences, and Sustainable Energy Assessment and Technologies Journals. His research interests include Data Analytics, Deep Learning, and Communication Systems based applications.
\end{IEEEbiographynophoto}
\vskip -2\baselineskip plus -1fil
\vspace{-0.25cm}
\begin{IEEEbiographynophoto}{Kapal Dev} is Senior Researcher at MTU, Ireland and senior research associate at University of Johannesburg, South Africa. He is AE in Springer Wireless Networks, Elsevier Physical Communication, IET Quantum Communication, IET Networks, Topic Editor in MDPI Network. He is contributing as GE in Q1 journals; IEEE TII, TNSE, TGCN, Elsevier COMCOM and COMNET. He served(ing) as Lead chair in MobiCom 2021, Globecom2021, $\&$ CCNC 2021 workshops. He contributed as PI for Erasmus + ICM, CBHE, and H2020 Co-Fund projects. His research interests include Blockchain, Wireless Networks and Artificial Intelligence. 
\end{IEEEbiographynophoto}
\vskip -2\baselineskip plus -1fil
\vspace{-0.25cm}
\begin{IEEEbiographynophoto}{Parus Khowaja} is pursuing her Ph.D. degree in financial analytics from University of Sindh, Jamshoro. She is currently working as an Assistant Professor at University of Sindh, Jamshoro. Her interests include Data analytics, Machine learning for Ambient Intelligence, Stock Portfolios, and Financial securities.
\end{IEEEbiographynophoto}
\vskip -2\baselineskip plus -1fil
\vspace{-0.25cm}
\begin{IEEEbiographynophoto}{Paolo Bellavista} received MSc and PhD degrees in computer science engineering from the University of Bologna, Italy, where he is now a full professor of distributed and mobile systems. His research activities span from pervasive wireless computing to online big data processing under quality constraints, from edge cloud computing to middleware for Industry 4.0 applications. He serves on several Editorial Boards, including IEEE COMST (Associate EiC), ACM CSUR, and Elsevier JNCA and PMC. He is the scientific coordinator of the H2020 BigData project IoTwins.
\end{IEEEbiographynophoto}
\end{document}